\documentclass[11pt]{article}
\usepackage{eacl2017}
\usepackage{times}
\usepackage{url}
\usepackage{latexsym}
\usepackage{amsmath}
\usepackage{multirow}
\eaclfinalcopy 
\usepackage{graphicx}
\usepackage{comment}
\usepackage{bm}
\usepackage{amsthm}
\usepackage{amssymb}
\usepackage{color}
\usepackage{version}
\usepackage{natbib}
\usepackage[dvipdfmx]{hyperref}
\hypersetup{%
setpagesize=false,
 bookmarksnumbered=true,%
 bookmarksopen=true,%
 colorlinks=true,%
 linkcolor=blue,
 citecolor=blue,
}
\title{Cutting-off Redundant Repeating Generations \\
for Neural Abstractive Summarization
}

\author{Jun Suzuki \and Masaaki Nagata \\
 NTT Communication Science Laboratories, NTT Corporation\\
 2-4 Hikaridai, Seika-cho, Soraku-gun, Kyoto, 619-0237 Japan\\
 {\tt \{suzuki.jun, nagata.masaaki\}@lab.ntt.co.jp} 
}

\date{}
\begin{document}
\maketitle
\begin{abstract}
 This paper tackles the reduction of redundant repeating generation that is often observed in RNN-based encoder-decoder models.
 Our basic idea is to jointly estimate the upper-bound frequency of each target vocabulary in the encoder and control the output words based on the estimation in the decoder.
 Our method shows significant improvement over a strong RNN-based encoder-decoder baseline and achieved its best results on an abstractive summarization benchmark.
 \footnote{This is a draft version of EACL-2017}
\end{abstract}

\section{Introduction}\label{sec:introduction}
   The RNN-based encoder-decoder (EncDec) approach 
   has recently been providing significant progress in various natural language generation (NLG) tasks, {\it i.e.},
   machine translation (MT)~\cite{DBLP:conf/nips/SutskeverVL14,cho-EtAl:2014:EMNLP2014}
   and abstractive summarization (ABS)~\cite{rush-chopra-weston:2015:EMNLP}.
   Since a scheme in this approach can be interpreted as a conditional language model, it is suitable for NLG tasks.
   However, one potential weakness is that it sometimes repeatedly generates the same phrase (or word).

   This issue has been discussed in the neural MT (NMT) literature as a part of a {\it coverage problem}~\cite{tu-EtAl:2016:P16-1,mi-EtAl:2016:EMNLP2016}.
   Such repeating generation behavior can become more severe in some NLG tasks than in MT.
   The {\it very short} ABS task in DUC-2003 and 2004~\cite{Over:2007:DC:1284916.1285157} is a typical example
   because it requires the generation of a summary in a pre-defined limited output space, such as ten words or 75 bytes.
   Thus, the repeated output consumes precious limited output space.
   Unfortunately, the coverage approach cannot be directly applied to ABS tasks
   since they require us to optimally find salient ideas from the input in a {\it lossy compression} manner, and thus the summary (output) length hardly depends on the input length; an MT task is mainly {\it loss-less} generation and nearly one-to-one correspondence between input and output~\cite{DBLP:journals/corr/NallapatiXZ16}.

   From this background, this paper tackles this issue and proposes a method to overcome it in ABS tasks.
   The basic idea of our method is to jointly estimate the upper-bound frequency of each target vocabulary that can occur in a summary during the encoding process and exploit the estimation to control the output words in each decoding step.
   We refer to our additional component as a {\bf word-frequency estimation (WFE) sub-model}.
   The WFE sub-model explicitly manages how many times each word has been generated so far and might be generated in the future during the decoding process.
   Thus, we expect to decisively prohibit excessive generation.
   Finally, we evaluate the effectiveness of our method on well-studied ABS benchmark data provided by Rush et al.~\shortcite{rush-chopra-weston:2015:EMNLP}, and evaluated in~\cite{chopra-auli-rush:2016:N16-1,nallapati-EtAl:2016:CoNLL,kikuchi-EtAl:2016:EMNLP2016,takase-EtAl:2016:EMNLP2016,DBLP:journals/corr/AyanaSLS16,gulcehre-EtAl:2016:P16-1}.

\section{Baseline RNN-based EncDec Model}
   The baseline of our proposal is an RNN-based EncDec model with an attention mechanism~\cite{luong-pham-manning:2015:EMNLP}.
   In fact, this model has already been used as a strong baseline for ABS tasks~\cite{chopra-auli-rush:2016:N16-1,kikuchi-EtAl:2016:EMNLP2016} as well as in the NMT literature.
   More specifically,  as a case study we employ a {\it 2-layer bidirectional LSTM} encoder and a {\it 2-layer LSTM} decoder with a global attention~\cite{DBLP:journals/corr/BahdanauCB14}.
   We omit a detailed review of the descriptions due to space limitations.
   The following are the necessary parts for explaining our proposed method.

   Let ${\bm X}\!=\!({\bm x}_i)^{I}_{i=1}$ and ${\bm Y}\!=\!({\bm y}_j)^{J}_{j=1}$ be input and output sequences, respectively, where ${\bm x}_i$ and ${\bm y}_j$ are one-hot vectors, which correspond to the $i$-th word in the input and the $j$-th word in the output.
    Let ${\cal V}^{\rm t}$ denote the vocabulary (set of words) of output.
   For simplification, this paper uses the following four notation rules:
   \begin{enumerate}
    \item $({\bm x}_i)^{I}_{i=1}$ is a short notation for representing a list of (column) vectors, {\it i.e.}, $({\bm x}_1,\dots, {\bm x}_I)=({\bm x}_i)^{I}_{i=1}$.
    \item $\bm{v}(a,D)$ represents a $D$-dimensional (column) vector whose elements are all $a$, {\it i.e.}, $\bm{v}(1,3)=(1, 1, 1)^{\top}$.
    \item $\bm{x}[i]$ represents the $i$-th element of $\bm{x}$, {\it i.e.}, $\bm{x}=(0.1, 0.2, 0.3)^{\top}$, then $\bm{x}[2]=0.2$.
    \item $M\!=\!|{\cal V}^{t}|$ and, $m$ always denotes the index of output vocabulary, namely, $m\in\{1,\dots,M\}$, and $\bm{o}[m]$ represents the score of the $m$-th word in ${\cal V}^{t}$, where $\bm{o}\in\mathbb{R}^{M}$.
   \end{enumerate}

   {\bf Encoder}: Let $\Omega^{\rm s}(\cdot)$ denote the overall process of our 2-layer bidirectional LSTM encoder.
   The encoder receives input $\bm{X}$ and returns a list of final hidden states $\bm{H}^{\rm s}=(\bm{h}^{\rm s}_i)^{I}_{i=1}$:
\begin{align}
 \displaystyle
 \bm{H}^{\rm s}
 &=
 \Omega^{\rm s}(\bm{X})
.
 \label{eq:encoder_process}
\end{align}
%

\begin{figure}[t]
 \centering
 \tabcolsep=1pt
 \small
 \begin{tabular}{rp{71mm}}
  \hline
  \ &\hspace{-1.2em}\textbf{Input}:
  $\bm{H}^{\rm s} = (\bm{h}^{\rm s}_i)^{I}_{i=1}$
  \hspace{\fill} {$\triangleright$ {\scriptsize list of hidden states generated by encoder}}
  \\
  \ &\hspace{-1.2em}\textbf{Initialize}:
      $s \leftarrow 0$
      \hspace{\fill}$\triangleright$ {\scriptsize $s$: cumulative log-likelihood}
      \\
  &\hspace{3.0em}
      $\hat{\bm{Y}} \leftarrow `BOS'$
      \hspace{\fill}$\triangleright$ {\scriptsize $\hat{\bm{Y}}$: list of generated words}
      \\
  &\hspace{3.0em}
      $\bm{H}^{\rm t} \leftarrow \bm{H}^{\rm s}$
      \hspace{\fill}$\triangleright$ {\scriptsize $\bm{H}^{\rm t}$: hidden states to process decoder}
      \\
  1:&\hspace{0.0em}
      $h \leftarrow (s, \hat{\bm{Y}}, \bm{H}^{\rm t})$
      \hspace{\fill}$\triangleright$ {\scriptsize triplet of (minimal) info for decoding process}
      \\
  2:&\hspace{0.0em}
      ${\cal Q}_{\tt w} \leftarrow {\tt push}({\cal Q}_{\tt w},h)$
      \hspace{\fill}$\triangleright$ {\scriptsize set initial triplet $h$ to priority queue ${\cal Q}_{\tt w}$}
      \\
  3:&\hspace{0.0em}
      ${\cal Q}_{\tt c} \leftarrow \{\}$
      \hspace{\fill}$\triangleright$ {\scriptsize prepare queue to store complete sentences}
      \\
  4:&\hspace{0.0em}
      \textbf{Repeat} \\
  5:&\hspace{3.0mm}
      $\tilde{\bm{O}} \leftarrow () $
      \hspace{\fill}$\triangleright$ {\scriptsize prepare empty list}
      \\
  6:&\hspace{3.0mm}
      \textbf{Repeat} \\
  7:&\hspace{6.0mm}
      $h \leftarrow {\tt pop}({\cal Q}_{\tt w})$
      \hspace{\fill}$\triangleright$ {\scriptsize pop a candidate history}
      \\
  8:&\hspace{6.0mm}
      $\tilde{\bm{o}} \leftarrow {\tt calcLL}(h)$
      \hspace{\fill}$\triangleright$ {\scriptsize see Eq.~\ref{eq:calc_output_score} }
      \\
  9:&\hspace{6.0mm}
      $\tilde{\bm{O}} \leftarrow {\tt append}(\tilde{\bm{O}},\tilde{\bm{o}})$
      \hspace{\fill}$\triangleright$ {\scriptsize append likelihood vector}
      \\
  10:&\hspace{3.0mm}
      \textbf{Until} ${\cal Q}_{\tt w} = \emptyset$
      \hspace{\fill}$\triangleright$ {\scriptsize repeat until ${\cal Q}_{\tt w}$ is empty}
      \\
  11:&\hspace{3.0mm}
      $\{(\hat{m},\hat{k})_{z}\}^{K-C}_{z=1} \leftarrow {\tt findKBest}(\tilde{\bm{O}})$
      \\
  12:&\hspace{3.0mm}
      $\{h_{z}\}^{K-C}_{z=1} \leftarrow {\tt makeTriplet}(\{(\hat{m},\hat{k})_{z}\}_{z=1}^{K-C} ) $
      \\
  13:&\hspace{3.0mm}
      ${\cal Q}' \leftarrow {\tt selectTopK}({\cal Q}_{\tt c}, \{h_{z}\}_{z=1}^{K-C}) $
      \\
  14:&\hspace{3.0mm}
      $({\cal Q}_{\tt w}, {\cal Q}_{\tt c}) \leftarrow {\tt SepComp}({\cal Q}') $
      \hspace{\fill}$\triangleright$ {\scriptsize separate ${\cal Q}'$ into ${\cal Q}_{\tt c}$ or ${\cal Q}_{\tt w}$}
      \\
  15:&\hspace{0.0em}
      \textbf{Until} ${\cal Q}_{\tt w} = \emptyset$
      \hspace{\fill}$\triangleright$ {\scriptsize finish if ${\cal Q}_{\tt w}$ is empty}
      \\
  \  &\hspace{-1.2em}\textbf{Output}: ${\cal Q}_{\tt c}$
      \\
  \hline
 \end{tabular}
 \caption{Algorithm for a $K$-best beam search decoding typically used in EncDec approach.}
 \label{fig:beam_search}
\end{figure}

   {\bf Decoder}:
   We employ a $K$-best beam-search decoder to find the (approximated) best output $\hat{\bm{Y}}$ given input $\bm{X}$.
   Figure~\ref{fig:beam_search} shows a typical $K$-best beam search algorithm used in the decoder of EncDec approach. 
   We define the (minimal) required information $h$ shown in Figure~\ref{fig:beam_search} for the $j$-th decoding process is the following triplet, $h=(s_{j-1}, \hat{\bm{Y}}_{j-1},  \bm{H}^{\rm t}_{j-1})$, where
   $s_{j-1}$ is the cumulative log-likelihood from step 0 to $j-1$,
   $\hat{\bm{Y}}_{j-1}$ is a (candidate of) output word sequence generated so far from step 0 to $j-1$, that is, $\hat{\bm{Y}}_{j-1}=(\bm{y}_0, \dots, \bm{y}_{j-1})$
   and
   $\bm{H}^{\rm t}_{j-1}$ is the all the hidden states for calculating the $j$-th decoding process.
   Then, the function {\tt calcLL} in Line 8 can be written as follows:
\begin{align}
 \displaystyle
 \tilde{\bm{o}}_j
 &= \displaystyle
 \bm{v}\big(s_{j-1}, {M}\big)
 +
 \log
 \big(
 {\tt Softmax} ( \bm{o}_{j} )
 \big)
 \nonumber\\
 \displaystyle
 \bm{o}_j
 &=
 \Omega^{\rm t}
 \big(
 \bm{H}^{\rm s}, \bm{H}^{{\rm t}}_{j-1}, \hat{\bm{y}}_{j-1}
 \big)
 ,
 \label{eq:calc_output_score}
\end{align}
   where ${\tt Softmax}(\cdot)$ is the {\it softmax} function for a given vector
   and $\Omega^{\rm t}(\cdot)$ represents the overall process of a single decoding step. 

   Moreover, $\tilde{\bm{O}}$ in Line 11 is a $({M\times (K-C)})$-matrix, where $C$ is the number of complete sentences in ${\cal Q}_{\tt c}$.
   The $(m,k)$-element of $\tilde{\bm{O}}$ represents a likelihood of the $m$-th word, namely $\tilde{\bm{o}}_j[m]$, that is calculated using the $k$-th candidate in ${\cal Q}_{\tt w}$ at the $(j-1)$-th step.
   In Line 12, the function {\tt makeTriplet} constructs a set of triplets based on the information of index $(\hat{m},\hat{k})$.
   Then, in Line 13, the function {\tt selectTopK} selects the top-$K$ candidates from union of a set of generated triplets at current step $\{h_{z}\}^{K-C}_{z=1}$and a set of triplets of complete sentences in ${\cal Q}_{\tt c}$.
   Finally, the function {\tt sepComp} in Line 13 divides a set of triplets ${\cal Q}'$ in two distinct sets whether they are complete sentences, ${\cal Q}_{\tt c}$, or not, ${\cal Q}_{\tt w}$.
   If the elements in ${\cal Q}'$ are all complete sentences, namely, ${\cal Q}_{\tt c} = {\cal Q}'$ and ${\cal Q}_{\tt w}=\emptyset$, then the algorithm stops according to the evaluation of Line 15.

\section{Word Frequency Estimation}
   This section describes our proposed method, which roughly
   consists of two parts:
   \begin{enumerate}
    \item a sub-model that estimates the upper-bound frequencies of the target vocabulary words in the output, and 
    \item  architecture for controlling the output words in the decoder using estimations.
   \end{enumerate}

\subsection{Definition}
   %
   Let $\hat{\bm{a}}$ denote a vector representation of the frequency estimation. 
   $\odot$ denotes element-wise product.
   $\hat{\bm{a}}$ is calculated by:
\begin{align}
 &
 \hat{\bm{a}}
 =
 \hat{\bm{r}}
 \odot
 \hat{\bm{g}}
 \nonumber \\
 &
 \hat{\bm{r}} = {\tt ReLU} (\bm{r})%
 ,
 \,\,\,\,\,\,
 \hat{\bm{g}} = {\tt Sigmoid} (\bm{g})
 \label{eq:wae_model}
 ,
\end{align}
   where ${\tt Sigmoid}(\cdot)$ and ${\tt ReLu}(\cdot)$ represent the element-wise sigmoid and ReLU~\cite{AISTATS2011_GlorotBB11}, respectively.
   Thus, $\hat{\bm{r}}\!\in\![0, +\infty]^{M}$,
   $\hat{\bm{g}}\!\in\![0,1]^{M}$, and
   $\hat{\bm{a}}\!\in\! [0, +\infty]^{M}$.

   We incorporate two separated components, $\hat{\bm{r}}$ and $\hat{\bm{g}}$, to improve the frequency fitting.
   The purpose of $\hat{\bm{g}}$ is to distinguish whether the target words occur or not, regardless of their frequency.
   Thus, $\hat{\bm g}$ can be interpreted as a {\it gate} function that resembles estimating the fertility in the coverage~\cite{tu-EtAl:2016:P16-1} and a switch probability in the copy mechanism~\cite{gulcehre-EtAl:2016:P16-1}.
   These ideas originated from such gated recurrent networks as LSTM~\cite{Hochreiter:1997:LSM:1246443.1246450} and GRU~\cite{DBLP:journals/corr/ChungGCB14}. 
   Then, $\hat{\bm{r}}$ can much focus on to model frequency equal to or larger than 1.
   This separation can be expected since $\hat{\bm{r}}[m]$ has no influence if $\hat{\bm{g}}[m]\!=\!0$.
   %

\subsection{Effective usage}
   %
   The technical challenge of our method is effectively leveraging WFE $\hat{\bm a}$. 
   Among several possible choices, we selected to integrate it as prior knowledge in the decoder.
   To do so, we re-define $\tilde{\bm{o}}_j$ in Eq.~\ref{eq:calc_output_score} as:
\begin{align}
 \displaystyle
 & \displaystyle
 \tilde{\bm{o}}_j
 = \displaystyle
 \bm{v}\big(s_{j-1}, {M}\big)
 +
 \log
 \left(
 {\tt Softmax}(\bm{o}_{j})
 \right)
 + \tilde{\bm{a}}_j
\nonumber
  .
\end{align}
   The difference is the additional term of $\tilde{\bm{a}}_j$,
   which is an {\it adjusted} likelihood for the $j$-th step originally calculated from $\hat{\bm{a}}$.
   We define $\tilde{\bm{a}}_j$ as: 
\begin{align}
 &
  \tilde{\bm{a}}_j
 =
 \log
 \left(
 {\tt ClipReLU}_{1}(\tilde{\bm r}_{j})
 \odot
 \hat{\bm{g}}
 \right)
 \label{eq:sub_model_usage}
 .
\end{align}
   $ {\tt ClipReLU}_{1}(\cdot)$ is a function that receives a vector and performs an element-wise calculation:
   $\bm{x}'[m]\!=\!\max\left(0, \min(1, \bm{x}[m]) \right)$ for all $m$ if it receives $\bm{x}$.
   We define the relation between $\tilde{\bm r}_j$ in Eq.~\ref{eq:sub_model_usage} and $\hat{\bm r}$ in Eq.~\ref{eq:wae_model} as follows:
\begin{align}
 \tilde{\bm r}_j
 &=\left\{
 \begin{array}{l}
  \hat{{\bm r}} \hspace{5.5em}\mbox{ if } j=1
   \\
  \tilde{\bm r}_{j-1} - \hat{\bm{y}}_{j-1} \hspace{1em}\mbox{ otherwise } 
 \end{array}
 \right.
 .
 \label{eq:manage_freq}
\end{align}
   Eq.~\ref{eq:manage_freq} is updated from $\tilde{\bm r}_{j-1}$ to $\tilde{\bm r}_{j}$ with the estimated output of previous step $\hat{\bm{y}}_{j-1}$.
   Since $\hat{\bm{y}}_{j} \!\in\! \{0,1\}^{M}$ for all $j$,
   all of the elements in $\tilde{\bm r}_j$ are monotonically non-increasing. 
   If $\tilde{\bm{r}}_{j'}[m] \!\leq\! 0$ at $j'$, then $\tilde{\bm{o}}_{j'}[m]\!=\!-\infty$ regardless of ${\bm{o}}[m]$.
   This means that the $m$-th word will never be selected any more at step $j'\leq j$ for all $j$.
   Thus, the interpretation of $\tilde{\bm r}_j$ is that it directly manages the upper-bound frequency of each target word that can occur in the current and future decoding time steps.
   As a result, decoding with our method never generates words that exceed the estimation $\hat{\bm r}$, and thus we expect to reduce the redundant repeating generation.

   Note here that our method never requires $\tilde{\bm{r}}_{j}[m] \!\leq\! 0$ (or $\tilde{\bm{r}}_{j}[m] \!=\! 0$) for all $m$ at the last decoding time step $j$, as is generally required in the coverage~\cite{tu-EtAl:2016:P16-1,mi-EtAl:2016:EMNLP2016,DBLP:journals/corr/WuSCLNMKCGMKSJL16}.
   This is why we say {\it upper-bound} frequency estimation, not just {\it (exact) frequency}.

\begin{figure}[t]
 \centering
 \tabcolsep=1pt
 \small
 \begin{tabular}{rp{71mm}}
  \hline
  \ &\hspace{-1.2em}\textbf{Input}:
  $\bm{H}^{\rm s} = (\bm{h}^{\rm s}_i)^{I}_{i=1}$
  \hspace{\fill}
  $\triangleright$ {\scriptsize list of hidden states generated by encoder}
  \\
  \ &\hspace{-1.2em}\textbf{Parameters}:
      ${\bm W}^{r}_1, {\bm W}^{g}_1 \!\!\in\! \mathbb{R}^{H \times H}$,
      ${\bm W}^{r}_2 \!\!\in\! \mathbb{R}^{M \times H}$,
      ${\bm W}^{g}_2 \!\!\in\! \mathbb{R}^{M \times 2H}$,
      \\
  1:&\hspace{0.0em}
      $\bm{H}^{r}_1 \leftarrow \bm{W}^{r}_1\bm{H}^{\rm s}$
      \hspace{\fill}
      $\triangleright$ {\scriptsize linear transformation for frequency model}
      \\
  2:&\hspace{0.0em}
      $\bm{h}^{r}_1 \leftarrow \bm{H}^{r}_1\bm{v}(1,M)$
      \hspace{\fill}
      $\triangleright$ {\scriptsize $\bm{h}^{r}_1\in\mathbb{R}^{H}$, $\bm{H}^{r}_1\in\mathbb{R}^{H\times I}$}
      \\
  3:&\hspace{0.0em}
      ${\bm{r}} \leftarrow \bm{W}^{r}_2 \bm{h}^{r}_1$
      \hspace{\fill}
      $\triangleright$ {\scriptsize frequency estimation}
      \\
  4:&\hspace{0.0em}
      $\bm{H}^{g}_1 \leftarrow \bm{W}^{g}_1\bm{H}^{\rm s}$
      \hspace{\fill}
      $\triangleright$ {\scriptsize linear transformation for occurrence model}
      \\
  5:&\hspace{0.0em}
      $\bm{h}^{g+}_2 \leftarrow \mbox{\tt RowMax}(\bm{H}^{g}_1)$
      \hspace{\fill}
      $\triangleright$ {\scriptsize $\bm{h}^{g+}_2\in\mathbb{R}^{H}$, and $\bm{H}^{g}_1\in\mathbb{R}^{H\times I}$}
      \\
  6:&\hspace{0.0em}
      $\bm{h}^{g-}_2 \leftarrow \mbox{\tt RowMin}(\bm{H}^{g}_1)$
      \hspace{\fill}
      $\triangleright$ {\scriptsize $\bm{h}^{g-}_2\in\mathbb{R}^{H}$, and $\bm{H}^{g}_1\in\mathbb{R}^{H\times I}$}
      \\
  7:&\hspace{0.0em}
      $ \bm{g} \leftarrow \bm{W}^{g}_2 \big( \mbox{\tt concat}(\bm{h}^{g+}_2, \bm{h}^{g-}_2)\big)$
      \hspace{\fill}
      $\triangleright$ {\scriptsize  occurrence estimation}
      \\
  \  &\hspace{-1.2em}\textbf{Output}: $({\bm{g}}, {\bm{r}})$
      \\
  \hline
 \end{tabular}
 \caption{Procedure for calculating the components of our WFE sub-model.}
 \label{fig:wfe_calculation}
\end{figure}

\subsection{Calculation}
   %
   Figure~\ref{fig:wfe_calculation} shows the detailed procedure for calculating ${\bm{g}}$ and ${\bm{r}}$ in Eq.~\ref{eq:wae_model}.
   For ${\bm r}$, we sum up all of the features of the input given by the encoder (Line 2) and estimate the frequency.
   In contrast, for ${\bm g}$, we expect Lines 5 and 6 to work as a kind of {\it voting} for both positive and negative directions
   since ${\bm g}$ needs just {\it occurrence} information, not {\it frequency}.
   For example, ${\bm g}$ may take large positive or negative values if a certain input word (feature) has a strong influence for occurring or not occurring specific target word(s) in the output.
   This idea is borrowed from the Max-pooling layer~\cite{DBLP:conf/icml/GoodfellowWMCB13}.

\subsection{Parameter estimation (Training)}
   %
   Given the training data, let $\bm{a}^{*}\in \mathbb{P}^{M}$ be a vector representation of the true frequency of the target words given the input, where $\mathbb{P}=\{0,1,\dots,+\infty\}$. 
   Clearly $\bm{a}^{*}$ can be obtained by counting the words in the corresponding output.
   We define loss function $ \Psi^{\rm wfe }$ for estimating our WFE sub-model as follows:
\begin{align}
 &
 \displaystyle
 \Psi^{\rm wfe } 
 \big(  
 {\bm X},
 \bm{a}^{*}, 
 {\cal W}
 \big)   
 = \displaystyle
 \bm{d}\cdot \bm{v}(1,M)
 \label{eq:wfe_loss}
 \\
 &
 \bm{d}
 =
 c_1 \max\big(\bm{v}(0,M), 
 \hat{\bm{a}}
 -
 \bm{a}^{*} 
 - \bm{v}(\epsilon,M)
 \big)^{b}
 \nonumber\\
 &
\,\,\,\,\,\,\, +
 c_2\max\big(\bm{v}(0,M), 
 \bm{a}^{*} 
 -
 \hat{\bm{a}}
 - \bm{v}(\epsilon,M)
 \big)^{b}
 ,
 \nonumber
\end{align}
   where $ {\cal W}$ represents the overall parameters.
   The form of $\Psi^{\rm wfe}(\cdot)$ is closely related to that used in support vector regression (SVR)~\cite{smola2004tutorial}.
   We allow estimation $\hat{\bm{a}}[m]$ for all $m$ to take a value in the range of $[\bm{a}^{*}[m]-\epsilon, \bm{a}^{*}[m]+\epsilon]$ with no penalty (the loss is zero).
   In our case, we select $\epsilon=0.25$ since all the elements of $\bm{a}^{*}$ are an integer.
   The remaining 0.25 for both the positive and negative sides denotes the {\it margin} between every integer.
   We select $b=2$ to penalize larger for more distant error, and $c_1 \!<\! c_2$, {\it i.e.}, $c_1\!=\!0.2, c_2\!=\!1$, since we aim to obtain upper-bound estimation and to penalize the under-estimation below the true frequency $\bm{a}^*$.

   Finally, we minimize Eq.~\ref{eq:wfe_loss} with a standard negative log-likelihood objective function to estimate the baseline EncDec model.

\begin{table}[t]
 \centering
 \small
 \tabcolsep=4pt
 \begin{tabular}{ c | r || c | r}
  \hline 
  Source vocabulary         & \multicolumn{3}{r}{$\dagger$ 119,507} \\
  Target vocabulary         & \multicolumn{3}{r}{$\dagger$  68,887} \\
  Dim. of embedding $D$     & \multicolumn{3}{r}{200}   \\
  Dim. of hidden state $H$  & \multicolumn{3}{r}{400}   \\
  \hline 
  Encoder RNN unit           & \multicolumn{3}{l}{2-layer bi-LSTM} \\
  Decoder RNN unit           & \multicolumn{3}{l}{2-layer LSTM with attention}   \\
   \hline
  Optimizer                  & \multicolumn{3}{l}{Adam (first 5 epoch)}\\
  \                          & \multicolumn{3}{l}{+ SGD (remaining epoch) $\star$ }\\
  Initial learning rate      & \multicolumn{3}{l}{0.001 (Adam) / 0.01 (SGD) } \\
  Mini batch size            & \multicolumn{3}{l}{256 (shuffled at each epoch) } \\
  Gradient clipping          & \multicolumn{3}{l}{10 (Adam) / 5 (SGD)}  \\
  Stopping criterion         & \multicolumn{3}{l}{max 15 epoch w/ early stopping}  \\
  \                          & \multicolumn{3}{l}{based on the val. set}  \\
  Other opt. options         & \multicolumn{3}{l}{Dropout = 0.3}               \\
    \hline
 \end{tabular}
 \caption{Model and optimization configurations in our experiments. $\dagger$: including special BOS, EOS, and UNK symbols. $\star$: as suggested in~\cite{DBLP:journals/corr/WuSCLNMKCGMKSJL16} }
  \label{table:configuration}
\end{table}
\begin{table*}[t]
 \small
 \centering
  \tabcolsep=2pt
  \begin{tabular}{ l  l || r | r | r || r | r | r  }
   \hline
   \      &       & \multicolumn{3}{c||}{DUC-2004 (w/ 75-byte limit)}
                  & \multicolumn{3}{c}{Gigaword (w/o length limit)}\\
   Method & Beam & ROUGE-1(R) & ROUGE-2(R) & ROUGE-L(R)  &  ROUGE-1(F) & ROUGE-2(F) & ROUGE-L(F) \\
   \hline
   EncDec                       &$B\!\!=\!\!1$    &          29.23  &             8.71  &           25.27 &           33.99  &           16.06  &           31.63\\
   (baseline)                  &$B\!\!=\!\!5$    &          29.52  &             9.45  &           25.80 &  $\dagger$34.27  &  $\dagger$16.68  &  $\dagger$32.14\\
   \hspace{1mm}our impl.)       &$B\!\!=\!\!10$   &$\dagger$ 29.60  &$\dagger$    9.62  &$\dagger$  25.97 &           34.18  &           16.51  &           31.97\\
   \hline
   \bf EncDec+WFE               &$B\!\!=\!\!1$    &          31.92  &             9.36  &           27.22 &           36.21  &           16.87  &           33.55\\
   (proposed)                   &$B\!\!=\!\!5$    &$\star$\bf 32.28 &$\star$\bf  10.54  &$\star$\bf 27.80 &$\star$\bf 36.30  &$\star$\bf 17.31  &$\star$\bf 33.88\\
   \                            &$B\!\!=\!\!10$   &          31.70  &            10.34  &           27.48 &           36.08  &           17.23  &           33.73\\
   \hline
   \hline
   \multicolumn{2}{c|}{ (perf. gain from $\dagger$ to $\star$)}
                              &      +2.68 &  +0.92 &   +1.83           &   +2.03 &  +0.63 &   +1.78 \\
   \hline
  \end{tabular}
 \caption{Results on DUC-2004 and Gigaword data: ROUGE-$x$(R): recall-based ROUGE-$x$, ROUGE-$x$(F): F1-based ROUGE-$x$, where $x\in\{1,2,L\}$, respectively.}
 \label{table:result_duc2004}
\end{table*}
\begin{table*}[t]
 \small
 \centering
 \tabcolsep=2pt
 \begin{tabular}{l | r | r | r || r | r| r}
  \hline
   \             & \multicolumn{3}{c||}{DUC-2004 (w/ 75-byte limit)}
                  & \multicolumn{3}{c}{Gigaword (w/o length limit)}\\
  \hline
  Method & ROUGE-1(R) & ROUGE-2(R) & ROUGE-L(R) & ROUGE-1(F) & ROUGE-2(F) & ROUGE-L(F)\\
  \hline
  ABS \cite{rush-chopra-weston:2015:EMNLP}
                       & 26.55  & 7.06  & 22.05 & 30.88 & 12.22 & 27.77\\
  \hline
  RAS \cite{chopra-auli-rush:2016:N16-1}&
      28.97  & 8.26  & 24.06 & 33.78  & 15.97 & 31.15\\
  \hline
  BWL \cite{DBLP:journals/corr/NallapatiXZ16}\footnote{The same paper was published in CoNLL~\cite{nallapati-EtAl:2016:CoNLL}. However, the results are updated in arXiv version.}&
      28.35  & 9.46  & 24.59 & 32.67  & 15.59 & 30.64\\
  \,\,\,(words-lvt5k-1sent$\dagger$)  &
      28.61  & 9.42  &25.24  &35.30   &$\dagger$16.64 & 32.62 \\
  \hline
  MRT \cite{DBLP:journals/corr/AyanaSLS16}&
      $\dagger$30.41  & $\dagger$\bf 10.87  & $\dagger$26.79 & $\dagger$\bf 36.54  & 16.59 & $\dagger$33.44\\
  \hline
  {\bf EncDec+WFE} [This Paper] &\bf  32.28  & 10.54  &\bf 27.80
                                   & 36.30 &\bf 17.31  &\bf 33.88 \\
  \hline
   \hline
   \multicolumn{1}{c|}{ (perf. gain from $\dagger$)}
                              &      +1.87 &  -0.33 &   +1.01           &   -0.24 &  +0.72 &   +0.44 \\
   \hline
 \end{tabular}
 \caption{Results of current top systems: `*': previous best score for each evaluation. $\dagger$: using a larger vocab for both encoder and decoder, not strictly fair configuration with other results.}
 \label{table:topsystem}
\end{table*}
%
\begin{table}[t]
 \small
 \centering
 \tabcolsep=2pt
 \begin{tabular}{ l || r | r |r |r|r   }
  \hline 
  \  True $\bm{a}^*$ $\backslash$ Estimation $\hat{\bm{a}}$  &   0  & 1 &   2  & 3 & $4\geq$  \\
  \hline 
  1              &  7,014  & 7,064     &  1,784  & 16  & 4  \\
  \hline 
  2              &  51     &    95     &   60    & 0 & 0  \\
  \hline 
  $3 \geq$       &   2    &  4      &   1     & 0 & 0  \\
  \hline 
 \end{tabular}
 \caption{Confusion matrix of WFE on Gigaword data: only evaluated true frequency $\geq 1$.}
 \label{table:result_wordset}
\end{table}
\begin{figure}[t]
 \centering
 \tabcolsep=1pt
 \small
 \begin{tabular}{rp{71mm}}
  \hline
  G: & {\tt china success at youth world championship shows preparation for \#\#\#\# olympics}\\
  A: & {\tt china \underline{germany} \underline{germany} \underline{germany} \underline{germany} and \underline{germany} at world youth championship}\\
  B: & {\tt china faces germany at world youth championship}\\
  \hline
  G:& {\tt British and Spanish governments leave extradition of Pinochet to courts}\\
  A:& {\tt spain britain seek shelter from \underline{pinochet 's} \underline{pinochet} case over \underline{pinochet 's}}\\
  B:& {\tt \underline{spain} britain seek shelter over pinochet 's possible extradition from \underline{spain}}\\
  \hline
  G:& {\tt torn UNK : plum island juniper duo now just a lone tree}\\
  A:& {\tt \underline{black women} \underline{black women} \underline{black} in \underline{black} code}\\
  B:& {\tt in plum island of the ancient}\\
  \hline
 \end{tabular}
 \caption{Examples of generated summary. G: reference summary, A: baseline EncDec, and B: EncDec+WFE. (underlines indicate repeating phrases and words)}
 \label{fig:raw_generation}
\end{figure}
\section{Experiments}
\label{sec:experiments}
   We investigated the effectiveness of our method on ABS experiments, which were first performed by Rush et al.,~\shortcite{rush-chopra-weston:2015:EMNLP}.
   The data consist of approximately 3.8 million training, 400,000 validation and 400,000 test data, respectively\footnote{The data can be created by the data construction scripts in the author's code: {{https://github.com/facebook/NAMAS}}.}.
   Generally, 1951 test data, randomly extracted from the test data section, are used for evaluation\footnote{As previously described~\cite{chopra-auli-rush:2016:N16-1} we removed the ill-formed (empty) data for Gigaword.}.
   Additionally, DUC-2004 evaluation data~\cite{Over:2007:DC:1284916.1285157}\footnote{http://duc.nist.gov/duc2004/tasks.html} were also evaluated by the identical models trained on the above Gigaword data. 
    We strictly followed the instructions of the evaluation setting used in previous studies for a fair comparison.
   Table~\ref{table:configuration} summarizes the model configuration and the parameter estimation setting in our experiments.

\subsection{Main results: comparison with baseline}
   Table~\ref{table:result_duc2004} shows the results of the baseline EncDec and our proposed EncDec+WFE.
   Note that the DUC-2004 data was evaluated by recall-based ROUGE scores, while the Gigaword data was evaluated by F-score-based ROUGE, respectively.
   For a validity confirmation of our EncDec baseline, we also performed {\tt OpenNMT} tool\footnote{{{http://opennmt.net}}}.
    The results on Gigaword data with $B=5$ were, 33.65, 16.12, and 31.37 for ROUGE-1(F), ROUGE-2(F) and ROUGE-L(F), respectively, which were almost similar results (but slightly lower) with our implementation.
    This supports that our baseline worked well as a strong baseline.
   Clearly, EncDec+WFE significantly outperformed the strong EncDec baseline by a wide margin on the ROUGE scores. 
    Thus, we conclude that the WFE sub-model has a positive impact to gain the ABS performance
   since performance gains were derived only by the effect of incorporating our WFE sub-model.

\subsection{Comparison to current top systems}
   Table~\ref{table:topsystem} lists the current top system results.
   Our method EncDec+WFE successfully achieved the current best scores on most evaluations.
   This result also supports the effectiveness of incorporating our WFE sub-model.
   
   MRT~\cite{DBLP:journals/corr/AyanaSLS16} previously provided the best results.
   Note that its model structure is nearly identical to our baseline.
   On the contrary, MRT trained a model with a sequence-wise minimum risk estimation, while we trained all the models in our experiments with standard (point-wise) log-likelihood maximization.
   MRT essentially complements our method.
   We expect to further improve its performance by applying MRT for its training since recent progress of NMT has suggested leveraging a sequence-wise optimization technique for improving performance~\cite{wiseman-rush:2016:EMNLP2016,shen-EtAl:2016:P16-1}.
   We leave this as our future work.

\subsection{Generation examples}
   %
   Figure~\ref{fig:raw_generation} shows actual generation examples.
   Based on our motivation, we specifically selected the redundant repeating output that occurred in the baseline EncDec.
   It is clear that EncDec+WFE successfully reduced them.
   This observation offers further evidence of the effectiveness of our method in quality.

\subsection{Performance of the WFE sub-model}
   %
   To evaluate the WFE sub-model alone,
   Table~\ref{table:result_wordset} shows the confusion matrix of the frequency estimation.
   We quantized $\hat{\bm{a}}$ by $\lfloor \hat{\bm{a}}[m]+0.5\rfloor$ for all $m$, where 0.5 was derived from the margin in $\Psi^{\rm wfe}$.
   Unfortunately, the result looks not so well.
   There seems to exist an enough room to improve the estimation.
    However, we emphasize that it already has an enough power to improve the overall quality as shown in Table~\ref{table:result_duc2004} and Figure~\ref{fig:raw_generation}.
   We can expect to further gain the overall performance by improving the performance of the WFE sub-model.
   %

\section{Conclusion}
   This paper discussed the behavior of redundant repeating generation often observed in neural EncDec approaches.
   We proposed a method for reducing such redundancy by incorporating a sub-model that directly estimates and manages the frequency of each target vocabulary in the output.
   Experiments on ABS benchmark data showed the effectiveness of our method, EncDec+WFE, for both improving automatic evaluation performance and reducing the actual redundancy.
   Our method is suitable for {\it lossy compression} tasks such as image caption generation tasks.
   %



\end{document}